# Chemistry Integrated Language Model using Hierarchical Molecular Representation for Polymer Informatics

Jihun Ahn[1†], Gabriella Pasya Irianti[1†], Vikram Thapar[1*], Su-Mi Hur[1,2*]

[1]Department of Polymer Engineering, Graduate School, Chonnam National University, Gwangju, 61186, Republic of Korea
[2]School of Polymer Science and Engineering, Chonnam National University, Gwangju, 61186, Republic of Korea
*Correspondence : thapar.09@gmail.com, shur@jnu.ac.kr

(Abstract) Machine learning has transformed material discovery for inorganic compounds and small molecules, yet polymers remain largely inaccessible to these methods. While data scarcity is often cited as the primary bottleneck, we demonstrate that strategic molecular representations can overcome this limitation. We introduce CI-LLM (Chemically Informed Language Model), a framework combining HAPPY (Hierarchically Abstracted rePeat unit of PolYmer), which encodes chemical substructures as tokens, with numerical descriptors within transformer architectures. For property prediction, De³BERTa, our descriptor-enriched encoder, achieves 3.5x faster inference than SMILES-based models with improved accuracy ($R^2$ score gains of 0.9–4.1% across four properties), while providing interpretable structure-property insights at the subgroup level. For inverse design, our GPT-based generator produces polymers with targeted properties, achieving 100% scaffold retention and successful multi-property optimization for negatively correlated objectives. This comprehensive framework demonstrates both forward prediction and inverse design capabilities, showcasing how strategic molecular representation advances machine learning applications in polymer science.

## 1. INTRODUCTION

The advent of machine learning has revolutionized materials discovery. The success of these models hinges on three pillars: quality data, robust network architectures, and effective material representation schemes. Traditional models employ handcrafted numerical descriptors, such as Morgan fingerprints, RDKit descriptors, and Mordred descriptors that encode chemical and structural information based on domain expertise [1–3]. While these approaches leverage chemical intuition, they suffer from fundamental inflexibility: descriptor sets require manual curations for each task, and the resulting representations fail to generalize across chemically diverse systems [4, 5]. Rather than relying on predefined descriptors, modern approaches train neural networks to extract features directly from raw molecular input in the form of graph-based molecular connectivity or string representation of molecules.

Among various approaches, Transformers, the foundational network architecture of Large Language Models (LLMs), have fundamentally transformed how we process sequential information across scientific domains due to their ability to capture "grammar" and "context". This capability has led researchers to leverage Transformer as feature extractors, generating numerical embeddings that encode substantial material-related knowledge and relevant information [6, 7]. Transformers have enabled breakthrough applications from protein structure prediction (AlphaFold2 [8]) to drug discovery [9] and materials design [10]. Transformers require conversion of chemical structures into molecular string representations, with SMILES (Simplified Molecular Input Line Entry System), and related string notations [11–13] being the most commonly adopted formats. Decoder-only LLMs such as GPT (Generative Pre-trained Transformer) have therefore been employed to directly generate SMILES-like sequences, often conditioning on target properties by prepending property tokens at the beginning of the sequence[14, 15]. Although this approach improves controllability, it still struggles with long sequence lengths, structural complexity, validity, and diversity[16, 17].

Yet polymers remain largely inaccessible to these methods. Despite offering unparalleled tunability through choices of functional group selection and chain architectures, their configurational freedom generates a combinatorically explosive search space[18–20]. Their hierarchical complexity from repeat units to functional groups to long chains with varied sequence arrangements resists atomistic encodings effective for other materials. Moreover, curated polymer databases contain structures for most properties, orders of magnitude fewer than protein and small-molecule repositories. Consequently, applying data-hungry Transformers to polymer systems remains exceptionally challenging. The problem intensifies for inverse design, where models must generate chemically valid polymer structures that satisfy property constraints. While Transformer encoder-based models for polymer property prediction exist, such as polyBERT[21] and TransPolymer[22], Transformer decoder (GPT)-based models for polymer design remain virtually nonexistent.

This limitation stems partly from how polymers are represented. The dominant chemical string representation, SMILES, encodes molecules as character strings where individual symbols denote atoms (C, N, O), bonds (=, #), and grammatical constructs (parentheses for branching, digits for ring closures). Transformer models for small molecules such as SMILES-BERT, ChemBERTa, MolGPT adopt character-level tokenization, treating each SMILES symbol as an independent token[15, 23, 24]. This approach succeeds for small molecules where compact sequences enable transformers to learn from millions of examples. However, for polymers, atom-level tokenization produces sequences too long for effective learning, grammatical special characters (parentheses, ring closures) complicate validity constraints, and the models struggle to maintain consistency across multi-scale structural features (e.g., ensuring stereochemistry while modifying functional groups). Reinforcement learning optimization, successfully applied to small-molecule property targeting [25], exacerbates these issues—models trained on character-level polymer SMILES disperses functional groups across dozens of non-contiguous tokens, obscures hierarchical structure, and generates sequences that strain Transformer capacity, often collapse toward invalid structures or fail to explore diverse chemical regions [26, 27].

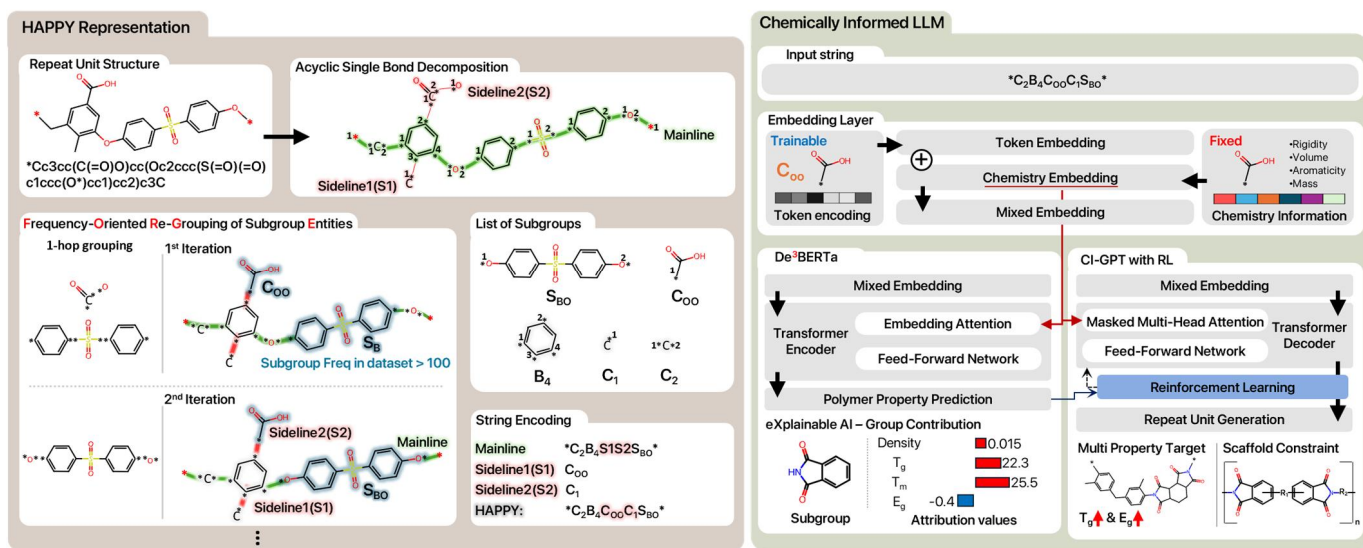

**Fig. 1 HAPPY representation construction and CI-LLM framework architecture for property prediction.** (Left) Schematic illustration of HAPPY representation construction. Repeat units undergo acyclic single-bond decomposition into mainline and sideline components. FORGE(Frequency-Oriented Re-Grouping of Entities) iteratively merges frequently recurring subgroups (>100 times) into larger subgroups, producing a compact string encoding. (Right) CI-LLM architecture for property prediction and generation. HAPPY tokens are combined with fixed chemical descriptors (rigidity, molecular refractivity, etc.) to form mixed embeddings. De³BERTa performs property prediction with explainable group-level attribution and CI-GPT CI-GPT with reinforcement learning enables targeted polymer generation including multi-property optimization and scaffold-constrained design.

We address these challenges through hierarchical coarse-graining into chemically meaningful motifs, mirroring both how chemists reason about molecules and the shift from character-level to sub-word tokenization in natural language processing[28]. We implement this through HAPPY (Hierarchically Abstracted rePeat unit of PolYmers), a representation that decomposes polymer structures into chemically meaningful subgroups [29], achieving significant sequence compression while simplifying syntax. Critically, we enrich these structural tokens with numerical chemical descriptors within a unified Transformer architecture, to our knowledge, the first such integration for polymer property prediction. This framework enables both interpretable property prediction and robust polymer generation via reinforcement learning, achieving the benefits of handcrafted features (data efficiency, interpretability) and learned representations (flexibility, task adaptation) while enabling interpretable, rational polymer design.

## 2. Result

### 2.1 HAPPY CI-LLM Framework

In our previous work, we introduced HAPPY (Hierarchically Abstracted rePeat unit of PolYmers), a string representation that decomposes polymer structures into chemically meaningful subgroups such as rings, functional groups, and recurring motifs. Building on this foundation, we developed a Chemically Informed LLM (CI-LLM) framework that integrates HAPPY with Transformer architectures for both property prediction and inverse design.

Starting from ~10,000 manually curated repeat units from PolyInfo[30], we applied the Frequency-Oriented Re-Grouping of Subgroup Entities(FORGE) algorithm (detailed in Section 4.1) to automatically identify recurring chemical motifs and constructure HAPPY representation as outline in Fig. 1a Each monomer is first converted to a canonical SMILES and fragmented by breaking acyclic single bonds, yielding initial subgroups. Subgroups are distinguished by both their local chemistry and connection point positions, enabling exact reconstruction of the original structure. The unique path between the two polymerizable ends defines the mainline, and all branches attached to mainline atoms are treated as sideline units (polymerizable ends highlighted in red in Fig. 1 left panel). We then apply our iterative subgroup-mining algorithm, FORGE, to merge frequently recurring local environments into larger subgroups. In each iteration, a subgroup and its nearest neighbors are merged into candidate larger subgroups (for example, the carboxylic and benzene–sulfonide subgroups in Fig. 1 left panel), whose frequencies are counted over the full dataset. Candidates occurring more than 100 times replace their constituent child subgroups, and this procedure is repeated until no new frequent subgroups are found. After convergence, every distinct subgroup is assigned an arbitrary token name. Using these token names, each repeat unit is encoded as a HAPPY string by traversing the mainline from one polymerizable end to the other. For every mainline token, any attached sideline subgroups are immediately written after the corresponding token.

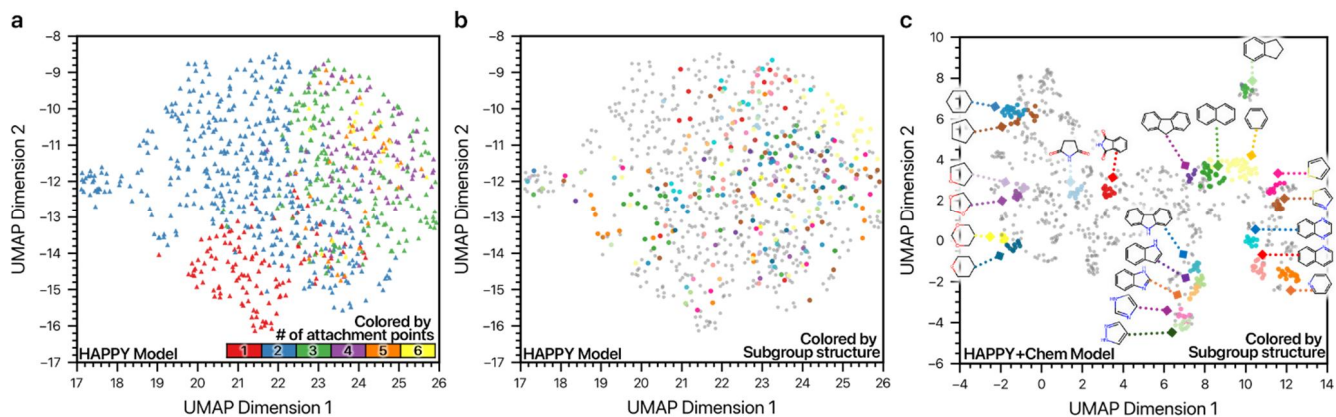

**Fig. 2 UMAP visualization of token embedding after pretraining (before property finetunning).** (a) HAPPY embeddings colored by number of attachement points, revealing a gradient from low (blue) to high (red) connectivity. (b) Same HAPPY embeddings colored by subgroup structure type. (c) HAPPY+Chem embeddings colored by structure type, with representative ring systems annotated. Chemical organization emerges from pretraining alone.

We use a DeBERTa-style Transformer encoder for all property-prediction models. For HAPPY, each monomer becomes a sequence of subgroup tokens along the mainline with interleaved sideline tokens. For SMILES, we adopt a standard tokenizer that breaks the string into tokens corresponding to atoms, ring indices, and bond markers. A special [CLS] token is prepended to each string. Each token is converted into a content embedding, encoding what that token is, and a position embedding, encoding where it appears relative to others. The attention layers use both embeddings, so each token's representation is updated based on both its identity and its location. The model is trained in two stages. In the pretraining stage, the model is trained with masked language modeling (MLM), where 15% of tokens are randomly corrupted and the model learns to recover the originals from context[31]. The [CLS] token is excluded from masking, allowing it to capture information from all other tokens throughout training. In the finetuning stage, this [CLS] vector is passed to a 1-layer neural network to predict a single polymer property. We train separate models for each of the four targets: glass-transition temperature, melting temperature, density, and bandgap. As illustrated in Fig. 1 right panel, our De$^3$BERTa(HAPPY+Chem) model augments each subgroup token with a fixed descriptor vector, encoding its local chemistry, obtained from RDKit and Mordred. The prepended [CLS] token receives a descriptor vector for the entire monomer. Descriptor embeddings are propagated through all attention layers, so interactions between tokens are influenced jointly by string context, relative position, and chemical features. As a comparison, we also include a Chem-only Transformer baseline that uses the same DeBERTa encoder but takes only descriptor-based embeddings as input and is trained from scratch without any MLM pretraining.

We use a GPT-style Transformer decoder with reinforcement learning (RL) to generate repeat unit structures with targeted properties. The GPT model works autoregressively so each token in a HAPPY or SMILES string attends only to itself and earlier tokens through masked multi-head self-attention. We first pretrain the model as a causal language model that predicts the next token from the previous context. The pretrained model is then evaluated by unconditional sampling of monomers and computing validity, novelty (fraction not in the training set), diversity, similarity to the training data, and Synthetic Accessibility score (SA score), with metric definitions given in Section 4.4. After this pretraining stage we apply policy-gradient reinforcement learning to steer generation toward design objectives. At each update the model samples 512 monomer strings and each sample receive a scalar reward that combines agreement with a single- or multi-property target, a diversity and similarity term that discourages near-duplicates and close copies of training molecules, and a chemical-realism term based on validity and SA score (reward function detailed in Section 4.5). The model parameters are updated to maximize the expected reward, so the generator moves toward desired objectives. We apply this GPT + RL framework to SMILES strings and to HAPPY strings augmented with chemical descriptors (CI-GPT) in direct analogy to the prediction models. Extensive details on the choice of descriptors including their incorporation in encoder and decoder models, and training setup are provided in Section 4.2 and 4.3 respectively.

## 2.2 UMAP Analysis of Pre-trained Subgroup Embeddings

Before evaluating downstream property prediction, we first assess what the model learns during pre-training. UMAP (Uniform Manifold Approximation and Projection) projections of subgroup embeddings highlight clear differences between HAPPY and HAPPY+Chem, indicating that the two representations organize chemical context in distinct ways (Fig. 2a–c).

In the HAPPY model, embeddings exhibit systematic topological structure (Fig. 2a,b). When colored by connectivity, the UMAP shows a pronounced gradient: subgroups separate by the number of attachment points, with mono- and di-substituted subgroups concentrated toward the lower-left and highly connected subgroups (5–6 attachment points) shifting toward the upper-right (Fig. 2a). This suggests the model internalizes a connectivity-driven hierarchy that reflects the architectural complexity of polymer substructures.

**Table 1 Summary of property prediction performance and representation characteristics.**

| | Dataset size | Pearson correlation coefficient(PCC) | Property Prediction Performance ($R^2$ score) | | | |
|---|---|---|---|---|---|---|
| | | | SMILES | HAPPY | Chemistry | HAPPY + Chemistry |
| Density | 1672 | 0.751 | 0.789± 0.066 | 0.735± 0.056 | 0.817± 0.030 | **0.830± 0.026** |
| Band gap energy | 3357 | 0.706 | 0.909 ± 0.011 | 0.870 ± 0.018 | 0.898 ± 0.016 | **0.913± 0.014** |
| Glass transition temperature | 6983 | 0.713 | 0.900± 0.009 | 0.900± 0.007 | 0.901± 0.007 | **0.909± 0.007** |
| Melting temperature | 3604 | 0.485 | 0.762± 0.018 | 0.758± 0.014 | 0.744± 0.026 | **0.773± 0.019** |
| Training time (second) | | | 11.9± 0.3 | 2.5± 0.1 | 2.6± 0.1 | 3.4± 0.1 |
| Sequence length | | | 55.2± 31.31 | 8.0± 2.73 | 8.0± 2.73 | 8.0± 2.73 |
| Vocabulary size | | | 59 | 865 | 865 | 865 |

When colored by structural class, clusters are only partially separated (Fig. 2b), implying that HAPPY embeddings capture some compositional similarity, but are dominated by topology.

Adding chemical descriptors reshapes the embedding space toward chemistry-aware organization (Fig. 2c). Ring-containing subgroups form distinct regions corresponding to ring families: phenyl subgroups concentrate in the upper area, fused polycyclic aromatics (e.g., naphthalene, anthracene) occupy an adjacent zone, and heteroaromatics (e.g., pyridine, thiophene) shift toward the lower-left, while saturated rings cluster separately. This demonstrates that chemical descriptors successfully enrich the representation space, adding fine- grained chemical distinction to HAPPY's topological foundation.

### 2.3 Property Predictions Performance

The effectiveness of our CI-LLM framework for property prediction is evaluated across four representative polymer properties: density ($\rho$), bandgap energy ($E_g$), glass transition temperature ($T_g$), and melting temperature ($T_m$). We compared four input representations: numerical chemical descriptors alone (Chem), HAPPY tokens, SMILES strings, and HAPPY enriched with chemical descriptors (HAPPY + Chem). Table 1 summarizes the $R^2$ score across all properties. HAPPY + Chem consistently achieves the highest prediction accuracy, outperforming SMILES by 4.1% for $\rho$, 0.4% for $E_g$, 0.9% for $T_g$, and 1.1% for $T_m$, while maintaining computational efficiency. Each alternative representation exhibits distinct limitations. HAPPY alone shows data sensitivity due to its extensive vocabulary of 865 tokens, showing poor performance with small datasets ($R^2$ = 0.735 ± 0.056 for $\rho$ with 1,672 datapoints) though matching SMILES when sufficient data are available ($R^2$ = 0.900 for $T_g$). The Chem-based model's effectiveness depends entirely on descriptor-property correlation, as evidenced by its lower performance on $T_m$ (Pearson correlation coefficient, PCC = 0.485) despite having more datapoints than $\rho$ (PCC = 0.751). SMILES, while achieving reasonable accuracy, suffers from computational inefficiency due to its long sequence length (55.2 ± 31.3 tokens), requiring 11.9 ± 0.3 seconds for 10,000 predictions, substantially longer than other representations. Importantly, we emphasis model performance not only by predictive accuracy but also by computational efficiency. Because HAPPY + Chem uses much shorter sequences (8.0 ± 2.73 tokens), it requires less memory and lower compute per inference, leading to faster throughput (3.4 ± 0.1 seconds for 10,000 predictions; ~3.5× faster than SMILES). This efficiency directly improves practical scalability and reduces energy consumption, facilitating more environmentally responsible AI while improving property prediction performance.

We further assessed data efficiency by systematically varying training set sizes from 100, 250, 500, 750, to 1000 datapoints (Fig. 3a-d). For each training set size, models were trained with multiple random splits (detailed in Section 4.3), and all resulting $R^2$ scores were averaged to obtain the final performance metrics. The corresponding standard deviations are represented by shaded regions of the same color

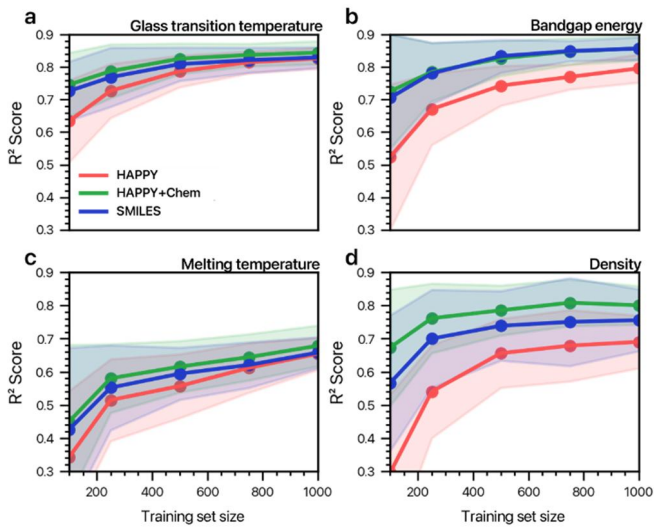

**Fig. 3 Data efficiency comparison across representations.** Prediction **accuracy** ($R^2$ score) versus training dataset size for four polymer properties (a) glass transition temperature, (b) bandgap energy, (c) melting temperature, and (d) density. HAPPY+Chem outperforms SMILES and HAPPY alone, particularly in low-data regimes. Shaded regions indicate standard deviations.

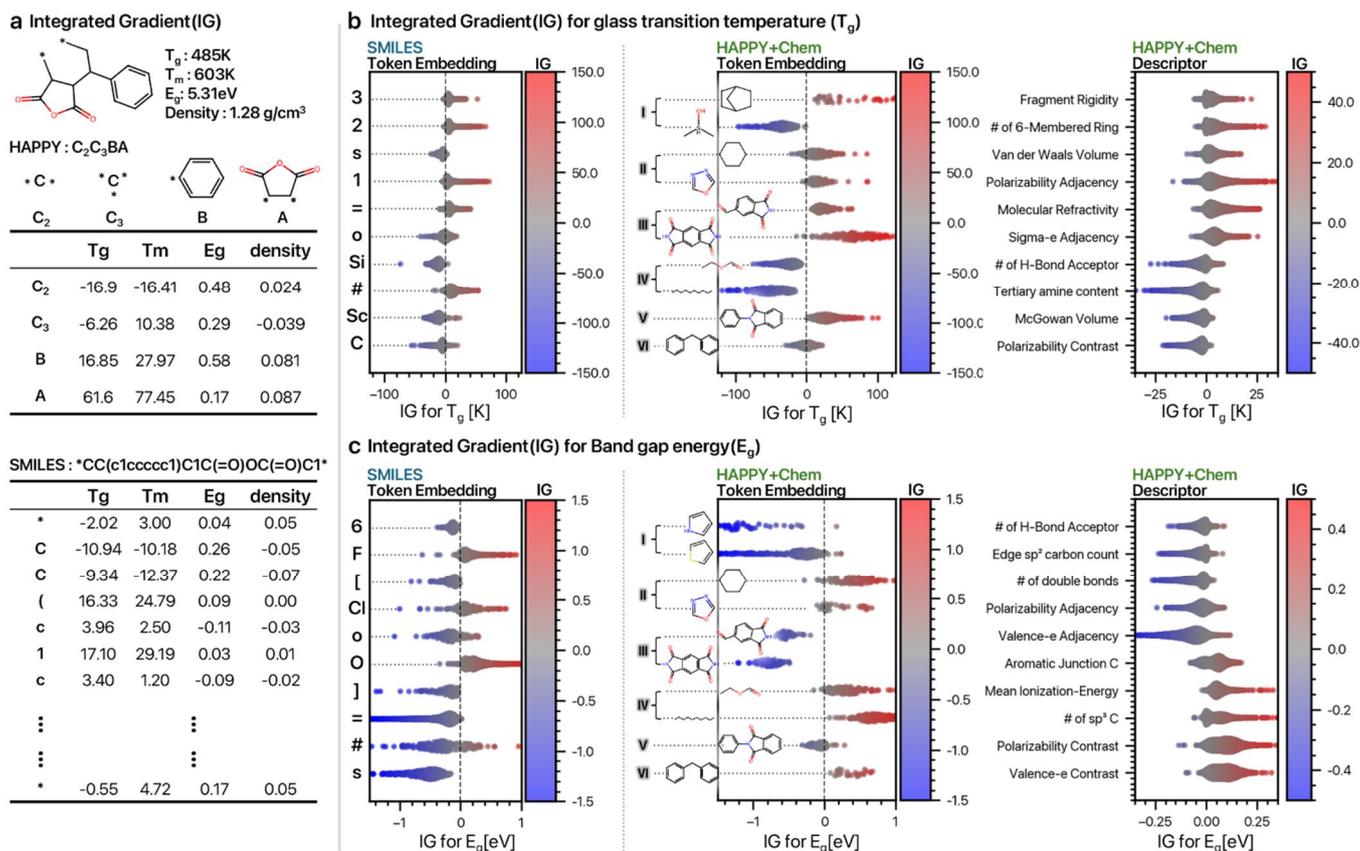

**Fig. 4 Integrated Gradients (IG) attribution analysis for interpretable structure-property relationships.** (a) Comparison of token-level IG attributions for a representative repeat unit, comparing SMILES-based DeBERTa and HAPPY+Chem De³BERTa models across four properties. (b-c) IG distributions across the full dataset for (b) $T_g$ and (c) $E_g$, showing SMILES characters (left), HAPPY+Chem subgroups with representative structures for Groups I–VI (center), and chemical descriptors contributions(right).

Overall, HAPPY alone performs poorly below 500 datapoints across all properties, aligning with our earlier interpretation on its expansive vocabulary requiring substantial training data. By contrast, the HAPPY + Chem model achieves higher $R^2$ than both HAPPY and SMILES even at the smallest training set size. With only 100 datapoints, HAPPY + Chem attains $R^2$ scores of 0.74± 0.10, 0.73± 0.19, 0.45± 0.23, and 0.68± 0.16 for $T_g$, $E_g$, $T_m$, and $\rho$, respectively, relative improvements of 1.4% ($T_g$), 2.7% ($E_g$), 2.6% ($T_m$), and 11.4% ($\rho$) compared to SMILES. This advantage persists across all training sizes, demonstrating that chemical descriptors effectively compensate for HAPPY's data requirements, making our CI-LLM framework particularly suitable for data-scarce polymer problems while maintaining computational efficiency.

### 2.4 Explainable AI: Token Level Attribution Analysis

We assessed interpretability using Integrated Gradients (IG), which assigns contribution scores to input tokens by accumulating gradients along a path from a baseline embedding to the actual input embedding (see Section 4.6). Fig. 4a compares IG attributions for the same repeat unit from SMILES- based DeBERTa and as HAPPY+ Chem De³BERTa models. In SMILES, attributions are distributed across many atom-level characters including non-chemical grammar symbols (parentheses, ring-closure digits), making it difficult to map scores onto chemically meaningful units. In contrast, HAPPY +Chem concentrates attribution on interpretable subgroups. For the representative example, subgroups corresponding to the phenyl ring and cyclic anhydride show strong positive contributions to $T_g$ and $T_m$, consistent with the expectation that rigid ring-containing subgroups.

Fig. 4b-c extends this analysis to the full dataset, comparing IG values for SMILES characters (left), HAPPY + Chem(center), and chemical descriptors (right). For $T_g$, SMILES characters linked to rings and unsaturation show a weakly positive tendency, but these symbols are shared across many distinct subgroups, so the signal remains diffuse. For $E_g$, SMILES exhibits more localized character-level effects, notably heteroatom-related characters.

The HAPPY+Chem subgroups are organized into groups based on their joint effects on $T_g$ and $E_g$. Group I shows the two strongest contributing subgroups for $T_g$ and $E_g$. Group II subgroups (e.g., cyclohexane and oxadiazole) increase both $T_g$ and $E_g$, consistent with increased rigidity and reduced/controlled electron delocalization [32–35]. Group III subgroups (e.g., fused imide-containing aromatics such as naphthalene diimide derivatives) increase $T_g$ while decreasing $E_g$, consistent with rigid fused rings that also expand π-conjugation and enhance delocalization. Conversely, Group IV subgroups are more aliphatic/$sp^3$-rich, lowering $T_g$ through flexibility while increasing $E_g$ due to reduced π-overlap and a larger σ–σ* gap. IG can also be computed for the chemical descriptors (Fig. 4b–c, right panel). For $T_g$, rigidity-related descriptors (e.g., fragment rigidity, number of 6-membered rings) contribute positively, and polarizability adjacency is also positive, consistent with stronger cohesive (dispersion) interactions. For $E_g$, descriptors related to $sp^2$ content, double bonds, and valence electron adjacency tend to decrease $E_g$ (enhanced π-delocalization), whereas ionization-energy-related descriptors and $sp^3$-carbon-related descriptors tend to increase $E_g$ by suppressing delocalization.

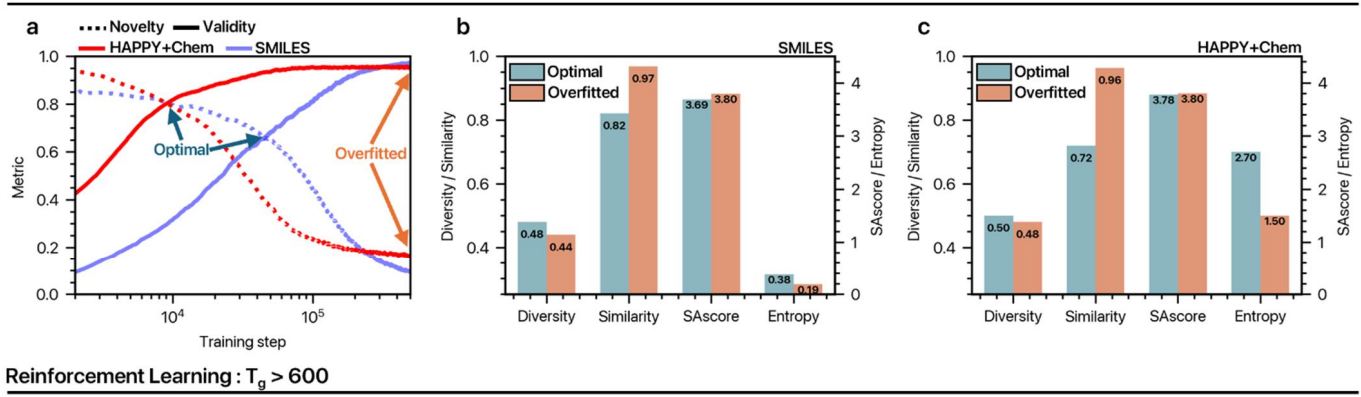

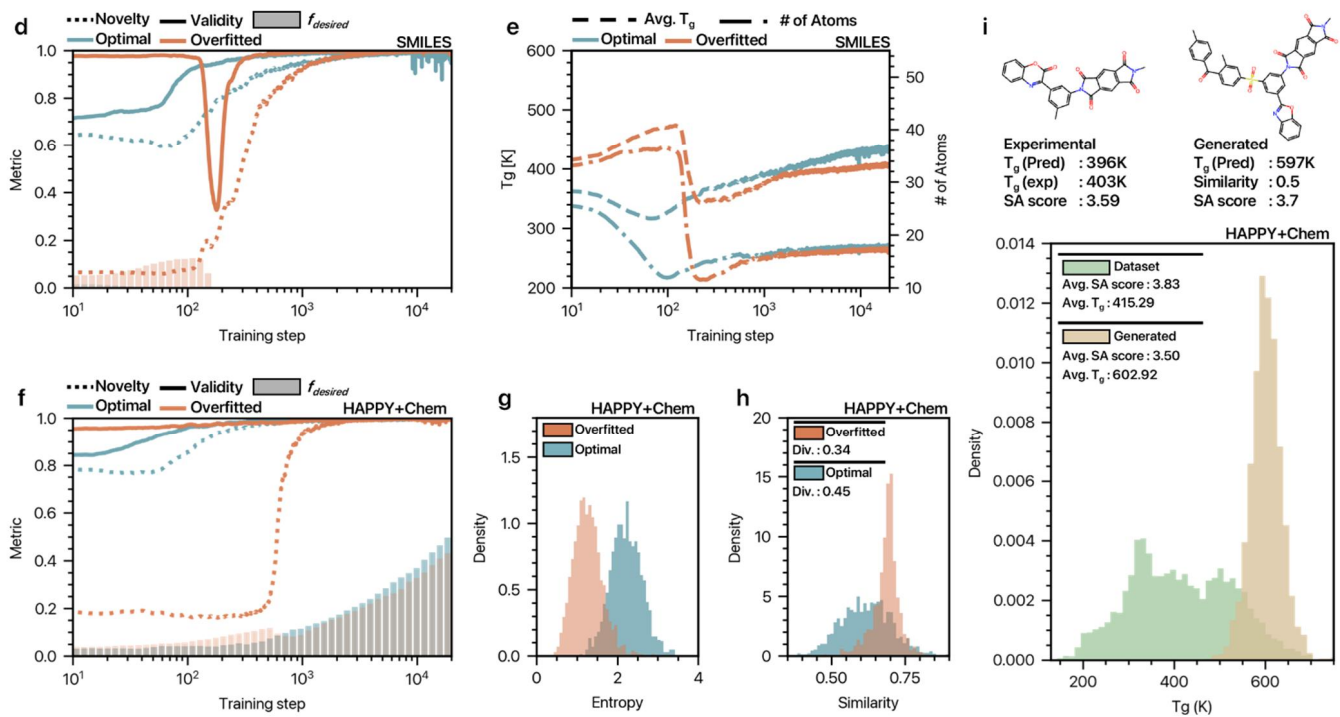

**Fig. 5 Unconditional generation and RL-guided high-$T_g$ repeat units optimization.** Top row: (a) Validity and novelty evolution during pretraining GPT for HAPPY+Chem and SMILES, with optimal and overfitted checkpoints indicated. (b-c) Diversity, similarity, SA score, and entropy metrics at both checkpoints for (b) SMILES and (c) HAPPY+Chem. Bottom row: RL-optimization targeting $T_g > 600$ K. SMILES models (d-e) fail entirely; validity collapses and molecules shrink to recover it, precluding high-$T_g$ generation. HAPPY+Chem models (f-h) successfully navigate the optimization, maintaining validity while reaching the target. (i) Generated $T_g$ distribution shifts substantially above training data, demonstrating effective exploration of high-$T_g$ chemical space.

This interpretability confirms that De³BERTa leverages chemically grounded features consistent with established structure-property relationships, while also highlighting subgroup-property associations that offer testable hypotheses for guiding synthesis.

### 2.5 Unconditional Generation and Reinforcement Learning-Guided Generation for High $T_g$ Repeat Units

To evaluate inverse design capabilities, we developed GPT-based generative models with reinforcement learning optimization. We trained autoregressive generators on both HAPPY + Chem and SMILES representations using next-token prediction, enabling the models to learn syntactic grammar and recurring chemistry patterns.

During pre-training, we periodically sampled 2,000 unconditional repeat units without target property or structural constraints and evaluated validity (fraction of chemically valid structure) and novelty (fraction absent from training set). Both models show increasing validity over training but declining novelty. At late training steps, both models overfit, with validity approaching 1.0 while novelty drops to ~0.1, indicating near-complete memorization of training examples. We refer to this late-stage checkpoint as the "overfitted" checkpoint.

We define the "optimal" checkpoint as the crossover between validity and novelty curves, where models still generate substantial fractions of both valid and novel structures. At this point, HAPPY + Chem achieves validity 0.83 and novelty 0.81, substantially outperforming SMILES (0.67 and 0.68). We further characterized the generated sets at the optimal and overfitted checkpoints using similarity (closeness to training structures), diversity (variation within the generated set), SA score (synthetic accessibility), and entropy (token-level exploration); see Section 4.4 for detailed definitions. Both representations maintain comparable diversity (≈0.45-0.50) and SA scores (3.7-3.8), confirming synthetically plausible structures. However, HAPPY + Chem exhibits lower similarity to training at the optimal point (0.72 vs 0.82 for SMILES) and much higher entropy (2.7 vs 0.38 at optimal point), indicating more exploratory generation enabled by its simpler grammar and larger vocabulary of chemically meaningful subgroups (Fig. 5a-c).

Having established these baseline characteristics, we coupled GPT with reinforcement learning to generate high-$T_g$ polymers. Starting from both the optimal and overfitted checkpoints identified above, we optimized for $T_g$ > 600K, training similarity ≈0.7, and SA score < 4.5. SMILES-based models fail to reach the target (Fig. 5d-e). The optimal SMILES model, starting with validity of ~0.7, produces no structures with $T_g$ > 600K throughout RL training, as the model prioritizes improving validity over generating high-$T_g$ candidates. The overfitted model, starting with near-perfect validity, initially generates some high-$T_g$ structures but these are largely memorized ones. As RL pushes exploration, validity temporarily drops while the model shifts toward simpler structures. Although validity eventually recovers, the resulting structures contain fewer than 20 atoms with $T_g$ well below 600 K. This demonstrates that under SMILES representation, RL struggles to maintain validity while generating structurally complex, high-$T_g$ candidates.

In contrast, HAPPY + Chem models successfully generate substantial fractions of high-$T_g$ structures (Fig. 5f-i). Both HAPPY + Chem models trained from optimal and overfitted checkpoints maintain high validity while increasing novelty, avoiding the validity drops observed for SMILES. However, the optimal model retains higher entropy, confirming more stochastic and diverse generation (Fig. 5g). The similarity distributions (Fig. 5h) show that the overfitted model's outputs concentrate narrowly around the similarity threshold of ~0.7, whereas the optimal model explores more broadly with lower similarity values, resulting in higher diversity (0.45 vs 0.34).

Fig. 5i compares the $T_g$ distributions of structures generated by the optimal HAPPY + Chem model from optimal checkpoint and training structures. Generated structures exhibit average $T_g$ of 603K with a narrow distribution tightly concentrated around the target, substantially higher than the training average of 415K. This demonstrates successful and precise evolution into sparsely populated high-$T_g$ regions of chemical space while maintaining low SA scores (~3.5). Representative examples (Fig. 5i, top) illustrate that the model generates novel structures with additional rings and branching compared to training examples, achieving predicted $T_g$ of ~597K with training similarity (0.5) and favorable SA scores (3.7).

## 2.6 Scaffold-Constrained and Multi-Objective Inverse Design with CI-GPT

Beyond single-property optimization, we explored two advanced scenarios. First, we extended the capability of CI-GPT to scaffold-constrained generation, enforcing presence of specific structural motifs. As a case study, we constrained generation to include bisphthalimide scaffolds. SMILES-based models fail catastrophically. The overfitted model can reach high scaffold fraction but collapses under RL pressure for novelty, while the optimal model reaches only ~10% before dropping to zero (Fig. 6a). In contrast, HAPPY-based models systematically increase scaffold fraction to 1.0 for both checkpoints, with the optimal model maintaining greater diversity.

Fig. 6b shows the optimal HAPPY model under a combined objective of scaffold constraint and high-$T_g$ (>600 K). The model ultimately attains average $T_g$ ~600K while driving the scaffold fraction to 1.0, progressively enriching the phthalimide scaffold while exploring chemically distinct high-$T_g$ derivatives. Representative structures at low and high $T_g$ stages are also shown in Fig. 6b; the high-$T_g$ structure (predicted $T_g$ of approximately 600K, SA score of 3.2) incorporates a pyridine unit in the phthalimidegroup, contributing to the elevated $T_g$ consistent with experimental trends[36].

Finally, we addressed multi-property optimization, maximizing two negatively correlated properties, $T_g$ and $E_g$ (Fig. 6c). These properties exhibit an inverse relationship with the target region ($T_g$ ≈ 600K and $E_g$ ≈ 4.5eV) sparsely represented in training data. Under strict constraints (similarity ≈ 0.7, SA score < 4.5), the model fails to reach the target region, as limited chemical space exploration prevents access to sparsely populated areas. Starting from the highly constraint model, removing the similarity constraint enables closer approach, while additional relaxing of the SA score threshold to 6.0 provides even more flexibility to reach the target (Fig. 6c-d). To validate the chemical rationality of the generated structures, we compared a representative output with an experimental reference from the training data ($T_g$ = 635 K, $E_g$ = 3.2 eV, containing a phthalimide group), shown in Fig. 6e-f. The CI-GPT-generated structure retains structural similarity to the experimental counterpart while incorporating targeted modifications: (1) the phthalimide group ($I_B$), which lowers $E_g$ due to extended π-conjugation across the fused aromatic-imide system, is replaced with $I_{BH}$, in which the saturated cyclohexane core interrupts electron delocalization, increasing $E_g$ while the rigid cyclic structure maintain high $T_g$, and (2) an additional cyclohexane unit is inserted into the backbone also contribute to raise $E_g$ while elevating $T_g$ (Fig. 4b-c). The resulting structure achieves $T_g$ of 600.20K and $E_g$ of 4.69eV with an SA score of 4.99, while maintaining a similarity of 0.4 to the experimental reference.

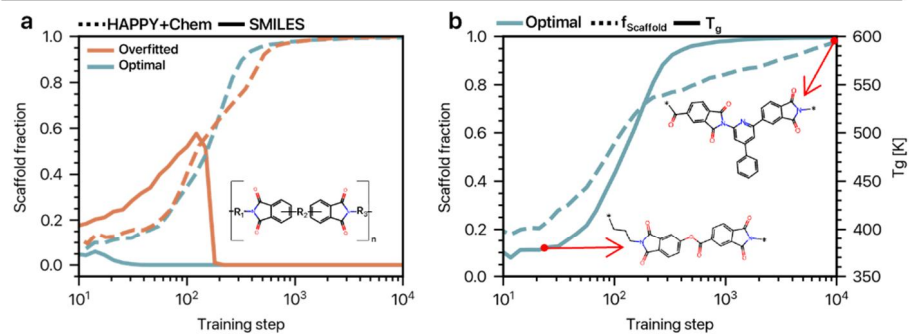
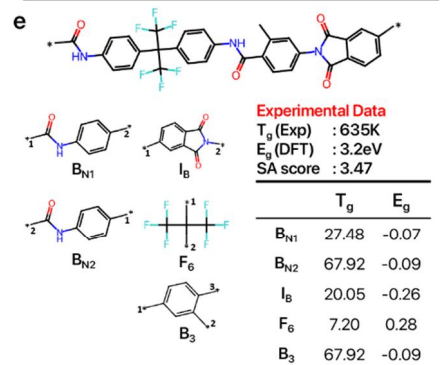
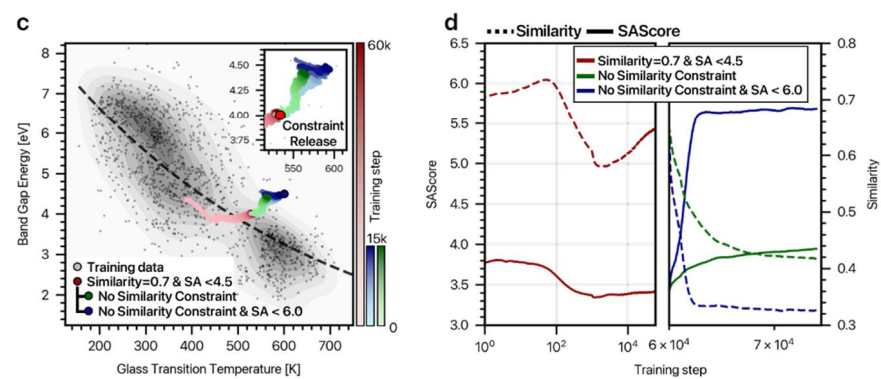
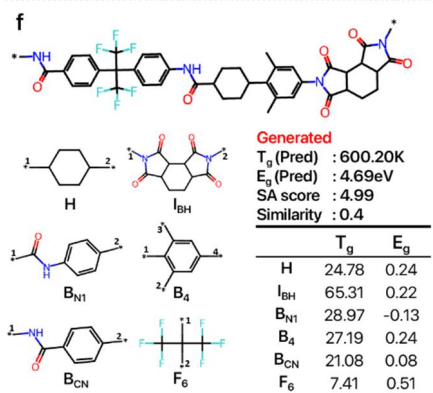

**Fig. 6 Scaffold-constrained and multi-objective inverse design with CI-GPT.** (a) Bis phthalimide scaffold fraction during RL training for SMILES and HAPPY+Chem models initialized from optimal and overfitted checkpoints. (b) Combined scaffold constraint and high-$T_g$ (>600K) optimization using the optimal HAPPY+Chem model, showing $T_g$ evolution, scaffold fraction, and representative structures at low and high $T_g$ stages. (c-d) Multi-objective optimization targeting $T_g \approx 600$K and $E_g \approx 4.5$eV, with representative generated structure compared to experimental reference shown with IG analysis(e-f).

## 3. CONCLUSION AND OUTLOOK

This work establishes molecular representation as a decisive factor in polymer informatics, not merely a preprocessing step, but a strategic choice that determines what models can learn and generate. By decomposing repeat units into chemically meaningful HAPPY subgroups and coupling them with chemical descriptors in the CI-LLM framework (De³BERTa and CI-GPT), we obtain a unified approach that supports accurate property prediction, physical interpretability, and controllable generative design. On the prediction side, De³BERTa delivers faster inference together with consistently higher accuracy than SMILES-based DeBERTa, while Integrated Gradients analysis provides subgroup-level attributions that directly link chemically intuitive motifs to model outputs—a level of interpretability unattainable with atom-level SMILES tokenization. On the generation side, CI-GPT combined with reinforcement learning solves progressively more demanding design tasks: it discovers high-$T_g$ polymers, enforces strict scaffold retention of a user defined subgroups, and, most notably, achieves simultaneous optimization of negatively correlated properties, producing polymers in sparsely populated regions of the training property space. Taken together, these results demonstrate that HAPPY and CI-LLMs provide a general scheme for rational polymer design, in which accurate prediction, interpretable motif attribution, scaffold control, and multi-objective optimization are treated within a single framework. While the present study focuses on homopolymer repeat units, the same representation and modeling principles can extend to copolymers and sequence-defined polymers.

The flexibility of the CI-LLM framework opens several avenues for future investigation. On the predictor side, our descriptor integration can also accommodate different families of descriptors (e.g., 3D, polymer physics–inspired, or processing-related descriptors), enabling targeted benchmarking of which descriptor sets are most informative for different classes of polymer properties. On the generator side, the success of reinforcement learning with CI-GPT opens the door to richer, user-defined design objectives. In addition to property targets and scaffold constraints, one can incorporate polymerization rules, synthetic and processing constraints, or application-specific performance metrics directly into the reward function, thereby steering the model toward more experimentally realizable high performing polymers.

## 4. ADDITIONAL DETAILS

### 4.1 Frequency-Oriented Re-Grouping of Subgroup Entities (FORGE)

As described in the overview of our framework, FORGE is an iterative subgroup-mining algorithm that merges frequently recurring local environments into larger subgroups. In each iteration, a subgroup and its nearest neighbors are merged to form candidate larger subgroups and the frequencies of all candidates are counted over the full dataset. Only candidates whose merged structure has at most two external attachment points are retained; subgroups that would introduce more than two attach points are discarded. Candidates that occur more than 100 times are promoted to new subgroups and replace their constituent child subgroups throughout the dataset. This creates a new set of HAPPY strings, on which the next iteration is run. If, after promotion and rewriting, the frequency of a previously promoted subgroups drops below the threshold, that subgroups is removed. The discover–merge–promote–rewrite–prune cycle is repeated until no new subgroup exceeds the frequency threshold.

During FORGE, different candidate subgroups can sometime overlap in a HAPPY string. To obtain a unique final tiling, we resolve this by using priority rules. We first prefer subgroups that were created in later FORGE iterations, and if two overlapping subgroups were created in the same iteration, we instead choose the one containing more atoms. These rules guarantee that every monomer is ultimately represented by a single, consistent set of FORGE-derived subgroups.

### 4.2 Incorporation of Chemical Information in De$^3$BERTa and CI-GPT Models

In the SMILES and HAPPY baseline DeBERTa-style models [37], each token is represented only by a learned content embedding that encodes which token it is. There is no explicit position embedding added to the input. Instead, DeBERTa handles relative position internally through its disentangled attention mechanism. Inside each self-attention block, the model combines the content representation of a token with an internal representation of its relative position to other tokens. The updated state at each position then reflects both what the neighboring tokens are and how they are arranged. In the De$^3$BERTa models we introduce an additional chemistry channel and use it in two ways. At the input level, every HAPPY subgroup token is assigned a fixed descriptor vector that summarizes its local chemistry, and the [CLS] token is assigned a descriptor vector for the whole monomer. These vectors are built from RDKit and Mordred 2D descriptors[38]. These descriptors are drawn from a library of roughly one thousand RDKit and Mordred 2D features, from which we select a subset with the strongest Pearson correlation to the target property so that the resulting descriptor vector is compatible with the model's embedding size. Before use, each selected descriptor is min–max scaled across the dataset to account for differences in numeric range. This chemistry vector is then simply added to the learned embedding at the input to make a content embedding, so each token starts with a single embedding that already mixes learned token information and fixed chemical information. Inside the attention blocks, content, chemistry, and relative-position signals are treated as separate channels whose pairwise interactions determine the attention weights. As a result, two tokens can attend strongly to one another because their learned representations are similar, because their descriptor patterns are similar, because they occupy a characteristic relative arrangement, or because several of these cues agree. As information flows through the stacked layers, each token representation becomes a blend of contextual information, relative arrangement, and chemical features. In the Chem-only baseline we remove the content channel entirely and feed only descriptor-based embeddings into the same backbone, giving a purely descriptor-driven reference model.

For the GPT-style models we use a standard Transformer decoder that generates tokens from left to right [39]. Each token has a learned embedding and an absolute position embedding, which are added together so the model knows both what the token is and where it appears in the string. In CI-GPT, every token also has a fixed chemistry embedding built from the same RDKit and Mordred descriptors, so the input to the model is the sum of three parts: learned, chemistry, and absolute position. In the SMILES-GPT baseline there is no chemistry term, so the input is just learned plus position. During self-attention, each token is allowed to attend only to itself and earlier tokens, which enforces left-to-right generation. The attention weights depend on both the content and chemistry channels, while the absolute position embedding provides order information, and there is no separate relative-position channel as in DeBERTa. As a result, in CI-GPT the next-token predictions depend jointly on the learned content representation, the fixed chemical descriptors, and the absolute positions of the tokens.

### 4.3 Training Configuration and Data Splits

All DeBERTa-style and GPT-style models used in this work share the same architecture with 3 Transformer layers, 256-dimensional token embeddings, and 8 attention heads. For DeBERTa pretraining, the dataset is split 80/20 into training and validation sets. We use masked language modeling (MLM) with the following scheme: 15% of tokens in each string are selected for possible corruption, of which 80% are replaced by a mask token, 10% by a random vocabulary token, and 10% are left unchanged. For both DeBERTa and GPT pretraining we use an initial learning rate of $5\times10^{-5}$, 5000 epochs, and a batch size of 64. DeBERTa pretraining uses a linear decay of the learning rate to 0.0, whereas GPT pretraining uses a cosine decay to 0.0. The DeBERTa property-prediction models are finetuned for 2500 epochs with the learning rate linearly decayed to half its initial value. For reinforcement learning with GPT we use a constant learning rate of $1\times10^{-5}$, with 512 generated monomers per update.

For each property and each data regime (100, 500, 750, 1000 points, and the full dataset), we evaluated prediction performance using repeated random splits with 5-fold cross-validation. We used 20 random splits for 100 datapoints, 10 splits for 500 and 750 datapoints, 5 splits for 1000 datapoints, and 2 splits for the full dataset. Each split was further partitioned into 5 folds, and models were trained and tested across all folds. All resulting $R^2$ scores were averaged to obtain the final performance for that property.

All experiments were conducted on a system equipped with an Intel Xeon Gold 6342 CPU (2.80 GHz) and an NVIDIA A100 GPU.

### 4.4 Evaluation Metrics for GPT Models

For each generative experiment we sample a fixed number of repeat units and evaluate several metrics. Validity is the fraction of generated strings that can be parsed into chemically valid molecules by RDKit. Novelty is the fraction of valid molecules that do not appear in the dataset. Similarity and internal diversity are computed from Tanimoto similarity between Morgan fingerprints [40] (radius 2, 1024 bits). For each generated monomer we define its similarity as the Tanimoto similarity to its nearest neighbor in the dataset. Its internal diversity contribution is the average of 1 -Tanimoto similarity over its 10 most similar neighbors within the generated set. The reported similarity and internal diversity are obtained by averaging these quantities over all generated valid monomers. Synthetic accessibility (SA) score[41] is computed using the standard RDKit implementation and ranges from 1 to 10, where 1 indicates molecules that are easy to synthesize and 10 indicates molecules that are very difficult to synthesize; we report the average SA score over valid molecules. Finally, entropy is measured as the Shannon entropy of the next-token probability distribution produced by GPT at each decoding step, averaged over tokens and sampled sequences, so that higher entropy indicates a broader and more exploratory output distribution.

### 4.5 Reward Function in Reinforcement Learning (RL)

During reinforcement learning, the reward is defined for each generated repeat units. If a generated string is invalid (RDKit cannot construct a molecule), its reward is set to zero. For valid repeat units, the reward is a weighted linear sum of several terms: a property term (or multi-property term), a diversity constraint, a similarity constraint, a specificity term, and a SA score constraint. Diversity and similarity are computed as in the GPT evaluation section from Tanimoto similarity of Morgan fingerprint. Specificity is designed to avoid overfitting to a single training molecule. For each generated repeat unit we first find its most similar repeat units in the dataset. We then count, over all $N$ generated repeat units in the batch, how many share the same nearest neighbor and define the specificity contribution as one minus this count divided by $N$. This quantity becomes small when many generated monomers collapse onto a single training example.

Each term is first computed in its "raw" form. For constraints where we want a quantity to match a target value, we use the corresponding metric value directly. For constraints of the form "greater than a threshold $X$" (such as properties that should be at least $X$), we apply a clamp: if the value is above $X$, we set the contribution to $X$; if it is below $X$, we keep the actual value. For constraints of the form "less than a threshold $X$", we use an analogous clamp on a negative scale: if the value is below $X$, we set the contribution to $-X$; if it exceeds $X$, the contribution becomes more negative as the violation increases.

At each reinforcement-learning update the model generates $N$ (with $N$ = 512) repeat units. For every term we therefore obtain $N$ raw values (one per repeat units). These values can lie on very different numerical ranges across terms, so for every term we apply min–max scaling over the $N$ samples to map them to a common range before forming the weighted sum. This ensures that the different terms contribute on comparable scales. All weights are set to 1 by default; in experiments where no similarity constraint is imposed, the weights for the similarity and specificity terms are set to 0 so that they do not affect the reward. Unless otherwise specified, we target SA score < 4.5, diversity > 0.6, specificity = 1, and similarity = 0.7 when defining the corresponding constraint terms.

### 4.6 Integrated Gradient (IG) Analysis

We assessed interpretability using Integrated Gradients (IG) [42], which assigns a contribution score to each input dimension by accumulating gradients along a smooth path from a baseline embedding to the actual input embedding. Intuitively, IG asks how the model's prediction changes as we "turn on" the input, starting from a neutral reference and moving step by step toward the true embedding. In practice, we approximate this by evaluating gradients at a finite number (200) of points between the baseline and the input and summing these contributions. We use an all-zero embedding as the baseline, so the resulting attributions indicate how much each embedding dimension contributes relative to an "absence of information" input.

For SMILES-based models, IG is computed with respect to the token embeddings of the input string. To obtain a single attribution per token, we aggregate the IG scores over all embedding dimensions belonging to that token. This yields a per-token importance score that indicates whether that SMILES token tends to increase or decrease the predicted property and by how much, on a relative scale. For HAPPY+Chem, we compute IG with respect to both the subgroup content embeddings and the chemistry descriptor embeddings. This produces two complementary types of attribution. First, we group IG scores by HAPPY subgroup token to identify which subgroups are most responsible for raising or lowering the predicted property. Second, we group IG scores by descriptor dimension to obtain an importance value for each chemical descriptor. Together, these analyses highlight not only which subgroups are most influential but also which underlying chemical features are most strongly associated with higher or lower values of the target properties.